\pdfoutput=1

\documentclass[11pt]{article}

\usepackage[preprint]{acl}

\usepackage{times}
\usepackage{latexsym}
\usepackage{amssymb}
\usepackage{booktabs}
\usepackage{amsmath}
\usepackage[T1]{fontenc}

\usepackage[utf8]{inputenc}

\usepackage{microtype}

\usepackage{inconsolata}

\usepackage{graphicx}
\usepackage{xspace}
\usepackage{enumitem}
\newcommand\method{PsyDraw\xspace}
\newcommand\high{High Consistency\xspace}
\newcommand\moderate{Moderate Consistency\xspace}
\newcommand\low{Low Consistency\xspace}
\newcommand\git{https://github.com/LYiHub/psydraw}

\setlist[itemize]{leftmargin=*}

\title{\method: A Multi-Agent Multimodal System for Mental Health \\ Screening in Left-Behind Children}

\author{
Yiqun~Zhang$^{1,2}$, Xiaocui~Yang$^{1}$, Xiaobai~Li$^{3}$, Siyuan~Yu$^{3}$, Yi~Luan$^{3}$,  \\
\textbf{Shi~Feng$^{1\dag}$, Daling~Wang$^{1}$, Yifei~Zhang$^{1}$} \\
$^1$Northeastern University, China, $^2$AI for Science, Shanghai AI Laboratory \\
$^3$Unitec Media Company Limited \\
\textbf{GitHub Repository:} \url{\git} \\
\textbf{Demo:} \url{https://udify.app/workflow/kHr98mSODQsaqArK}
}

\begin{document}
\maketitle
\begin{abstract}
Left-behind children (LBCs), numbering over 66 million in China, face severe mental health challenges due to parental migration for work. Early screening and identification of at-risk LBCs is crucial, yet challenging due to the severe shortage of mental health professionals, especially in rural areas. While the House-Tree-Person (HTP) test shows higher child participation rates, its requirement for expert interpretation limits its application in resource-scarce regions.
To address this challenge, we propose \textbf{\method}, a multi-agent system based on Multimodal Large Language Models that assists mental health professionals in analyzing HTP drawings. The system employs specialized agents for feature extraction and psychological interpretation, operating in two stages: comprehensive feature analysis and professional report generation.
Evaluation of HTP drawings from 290 primary school students reveals that 71.03\% of the analyzes achieved \textbf{\high} with professional evaluations, 26.21\% \moderate and only 2.41\% \low. The system identified 31.03\% of cases requiring professional attention, demonstrating its effectiveness as a preliminary screening tool. Currently deployed in pilot schools, \method shows promise in supporting mental health professionals, particularly in resource-limited areas, while maintaining high professional standards in psychological assessment.
\end{abstract}

\def\thefootnote{\dag}\footnotetext{Corresponding author.}\def\thefootnote{\arabic{footnote}}

\section{Introduction}
\textbf{Left-behind children (LBCs)}, resulting from parental economic migration, face severe mental health challenges \cite{WANG2020105135, Zhue013502, ijerph21060793}. Studies show that LBCs exhibit 50-80\% higher incidence of suicidal thoughts compared to non-LBC peers \cite{fellmethHealthImpactsParental2018, racaiteEmotionalBehaviouralProblems2024}, along with increased rates of anxiety and depression. In 2020, there are approximately 66.93 million LBCs in China \cite{StatusChinaMigrant}. 
This poses a significant challenge for China’s mental health system, which has approximately 50,000 clinical psychologists. In contrast, the United States has about 200,000, despite having only a quarter of China’s population \cite{fang2020mental}. 
The situation is particularly alarming as these limited mental health resources are concentrated in urban areas, while most LBCs are in underserved rural regions \cite{caiping2013number, cheng2020urban}.
Given the scale of the problem and the severe shortage of mental health professionals, implementing effective mental health screening becomes particularly challenging. 
To address this issue, various assessment methods have been explored. Traditional standardized questionnaires such as the Child Behavior Checklist \cite{achenbach1991child} and the Strengths and Difficulties Questionnaire \cite{goodman1997strengths} are commonly used, yet often face low compliance rates due to their lengthy and tedious nature \cite{palmerInvestigationClinicalUse2000a}.
As an alternative approach, projective tests like the \textbf{House-Tree-Person (HTP) test} \cite{burns1987kinetic}, which evaluates psychological state by analyzing children’s drawings of houses, trees, and persons, offers a more engaging experience through drawing activities, increasing children’s willingness to participate \cite{palmerInvestigationClinicalUse2000a}.

Despite its benefits, the HTP test lacks objective indicators and requires expert interpretation, challenging its use in areas with limited mental health resources. 
Interpreting HTP drawings and generating analysis reports is a complex task that requires professional knowledge of psychology and image understanding.
The emergence of multimodal large language models (MLLMs), with their ability to process and analyze images and text simultaneously, presents a new opportunity to address this challenge.  
MLLMs have demonstrated remarkable capabilities in visual understanding tasks such as Visual Question Answering (VQA) \cite{liu2023visualinstructiontuning, bai2023qwenvlversatilevisionlanguagemodel} and have shown promising applications in healthcare diagnosis \cite{li2024llava}, suggesting their potential for automating HTP test interpretation.

To address this challenge, we propose \method, a multi-agent system based on MLLMs. \method aims to evaluate mental health status and identify positive and negative factors by analyzing HTP drawings. 
The system's workflow is divided into two main stages: \textbf{feature analysis} and \textbf{report generation}. Each stage is completed by multiple collaborating agents, with each agent guided by prompts crafted with professional knowledge to accomplish its specific tasks. To validate the effectiveness of \method, we analyze HTP drawings from 290 Chinese primary school students, generating corresponding mental health reports. These reports are subsequently evaluated and annotated by the subjects' psychological teachers. The results illustrate that 71.03\% of the analyses are rated as \textbf{\high}, 26.21\% as \textbf{\moderate}, and only 2.41\% as \textbf{\low}. 
Designed as an \textbf{assistive tool} for mental health professionals rather than a standalone solution, \method has been deployed in several schools to support psychological teachers in early screening. These preliminary findings demonstrate \ method's potential in automating HTP test analysis, offering an innovative solution to the shortage of mental health professionals in assessing the psychological well-being of LBCs.

\begin{figure*}[t]
    \centering
    \includegraphics[width=\linewidth]{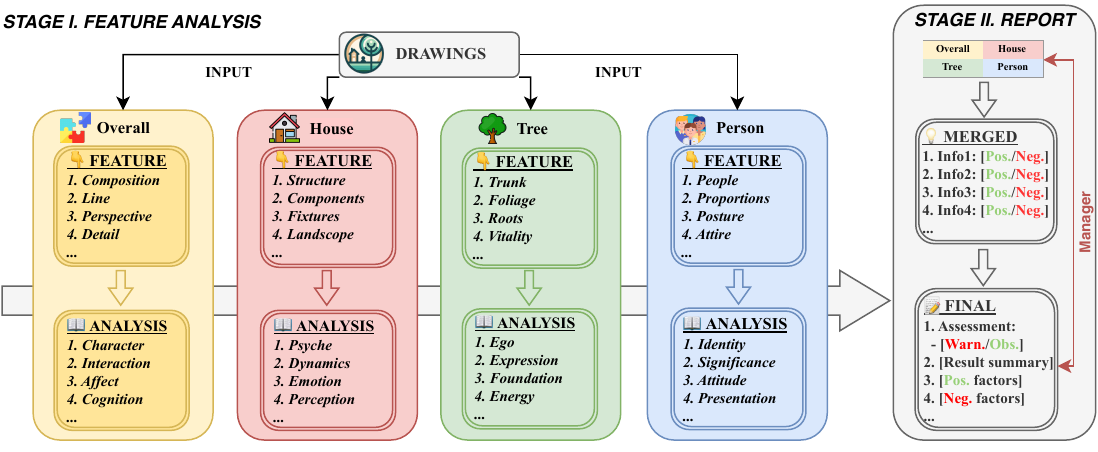}
    \caption{The workflow of \method.}
    \label{fig: workflow}
\end{figure*}

\section{House-Tree-Person Test}
The House-Tree-Person (HTP) Test is a projective psychological assessment tool designed to evaluate an individual's personality, emotional functioning, and interpersonal relationships through their drawings \cite{burns1987kinetic, palmerInvestigationClinicalUse2000a}. The test involves asking participants to produce separate drawings of a house, a tree, and a person, each of which is theorized to represent distinct facets of the individual's internal world \cite{marzolf1972house}. In particular, the house is commonly interpreted as symbolizing familial structures, perceived security, and relational dynamics within the home environment; its architectural features (e.g., size, solidity, presence of doors and windows) are posited to offer insights into feelings of belonging, openness, or vulnerability. The tree is frequently regarded as an emblem of the self's core strength, stability, and growth potential, with attributes such as trunk width, branch span, and leaf density informing evaluators about the client's sense of resilience, energy, and capacity for emotional regeneration. Finally, the person figure is often conceptualized as reflecting core aspects of identity, self-esteem, and attitudes toward others; variations in posture, facial expression, and detail may indicate internalized social expectations, degrees of self-confidence, and latent interpersonal tensions. 

As a flexible, relatively unstructured measure, the HTP has found applications in clinical, educational, and forensic settings \cite{soll2019use}, providing a rich, projective canvas for individuals who may struggle to articulate their inner experiences verbally. Its clinical utility encompasses identifying latent distress, gauging self-perception, and informing treatment approaches; in educational contexts, it assists in understanding a child's developmental status, affective challenges, and socialization patterns; and in forensic scenarios, it can complement other assessments to evaluate personality dynamics relevant to legal considerations \cite{guoAnalysisScreeningPredicting2023}. 
However, the complexity and subjectivity inherent in interpreting these drawings pose significant challenges for widespread application, particularly in resource-limited settings. While recent studies have attempted to automate HTP analysis using machine learning techniques \cite{salar2023artificial, de2024identifying}, these approaches primarily focus on extracting specific drawing features and cannot generate comprehensive psychological assessments. The emergence of MLLMs, with their advanced visual understanding and text generation capabilities, coupled with appropriate task decomposition and agent collaboration, presents a promising direction for more holistic and generalizable HTP analysis.

\section{Methodology}
\subsection{Task Definition}
The automated analysis of HTP test interpretation is formulated as a visual-linguistic understanding task. Given an HTP drawing $I$ as input, the task consists of three main components:

\paragraph{Feature Extraction}
Extract and analyze psychological indicators $\phi(I)$ from four aspects:\\
\textbf{Overall features}: spatial organization (e.g., perspective, proportion, placement), drawing characteristics (e.g., line quality, symmetry, transparency), and stylistic elements (e.g., detail level, ground lines, shading).\\
\textbf{House features}: architectural characteristics (e.g., size, structure, windows, doors) reflecting family dynamics and security perception.\\
\textbf{Tree features}: growth patterns (e.g., trunk structure, branch distribution, foliage density) indicating emotional stability and development.\\
\textbf{Person features}: figure attributes (e.g., posture, facial expression, proportions) revealing self-perception and social relationships.

\paragraph{Psychological Analysis}
Generate a comprehensive psychological assessment report $R = \psi(\phi(I))$, where $R$ contains a structured analysis of individual elements and overall composition, interprets identified characteristics' psychological implications and maintains professional terminology while ensuring readability.

\paragraph{Risk Classification}
Produce a binary risk classification $C = \tau(R) \in \{Warning, Observation\}$, where \textit{\textbf{Warning}} indicates potential psychological risks requiring professional attention. In contrast, \textit{\textbf{Observation}} suggests no significant concerns detected from the drawing.

This formulation supports mental health professionals in the early psychological screening of LBC by providing detailed analytical reports and preliminary risk assessments, helping reduce their workload while maintaining professional standards.

\subsection{PsyDraw}
HTP test interpretation requires comprehensive professional knowledge and typically involves analysis of over 100 distinct features. Such complexity makes it challenging for a single model to handle feature extraction and simultaneously generate professional, reliable reports. We propose PsyDraw, an automated system that draws upon professional HTP analysis manuals \cite{burns1987kinetic} and implements a multi-agent collaborative framework to address these challenges. This framework divides complex tasks into manageable units, mimicking how human experts collaborate in professional settings and consists of three types of agents: analysis agents (Overall, House, Tree, and Person) for detailed feature extraction and preliminary analysis, expert agents for report integration and psychological interpretation, and a manager agent that coordinates the workflow and handles exceptional cases. The analysis process is structured into two main stages: HTP image feature analysis (\textbf{\textit{Stage 1}}) and report generation (\textbf{\textit{Stage 2}}).

Figure \ref{fig: workflow} illustrates the workflow of PsyDraw. In \textbf{\textit{Stage 1}}, four analysis agents (\textbf{Overall}, \textbf{House}, \textbf{Tree}, and \textbf{Person}) work in parallel, each focusing on a different aspect of HTP image interpretation. Following a consistent two-step strategy, these agents extract specific features using specialized prompts and then analyze these features using professional prompts to generate preliminary reports. The \textbf{Overall} Agent focuses on depth, size, and position, revealing the subject's cognition, attitude, and emotional response to the environment. The \textbf{House} Agent examines components like roofs and windows, interpreting family background and kinship relationships. The \textbf{Tree} Agent evaluates growth patterns, assessing subconscious self-image and psychological maturity. The \textbf{Person} Agent analyzes human figures, examining self-image establishment and psychological defense mechanisms.

In \textbf{\textit{Stage 2}}, expert agents take over to integrate and finalize the analysis. These agents consolidate the preliminary reports from \textbf{\textit{Stage 1}}, label features with tendencies (positive, negative, neutral), and synthesize a comprehensive report. The final report includes an assessment result (\textbf{\textit{Warning}} or \textbf{\textit{Observation}}), an analysis summary, and detailed analyses of positive and negative factors.

Throughout both stages, the manager agent maintains system stability by handling three types of exceptional situations: (1) service rejection due to medical advice restrictions, which triggers automatic retry requests; (2) detection of severe harmful content in drawings (e.g., suicide, violence), which prompts immediate warning messages and professional assistance recommendations; and (3) technical issues such as network failures, which are resolved through retry mechanisms.

\section{Experimental}
\subsection{Experimental Setup}
This study is conducted in a Chinese primary school for left-behind children, accessed through a non-profit organization, collecting a total of \textbf{290} HTP drawings. The sample comprises 145 third-grade students and 145 fifth-grade students. During the drawing process, each participant is instructed on the basic HTP test rules: \textit{using a provided pen and paper to create a drawing containing a house, tree, and person, with no time limit}. In our experiments, the \method system utilizes Claude-3.5-sonnet\footnote{https://www.anthropic.com/news/claude-3-5-sonnet} as the text-only model and GPT-4o\footnote{https://openai.com/index/hello-gpt-4o/} as the multimodal model. We set the temperature parameter to 0.2 and Top \textit{P} to 0.75 for all models while keeping other parameters at their default values. All data collection procedures are conducted with explicit consent from teachers and students. All personal identifiers are removed from the drawings before analysis to ensure privacy protection. The development team of \method can only access the anonymized drawings, with no additional personal information or demographic data provided. This data anonymization protocol is implemented to maintain participant confidentiality while allowing for objective system evaluation.

\begin{figure}[t]
    \centering
    \includegraphics[width=\linewidth]{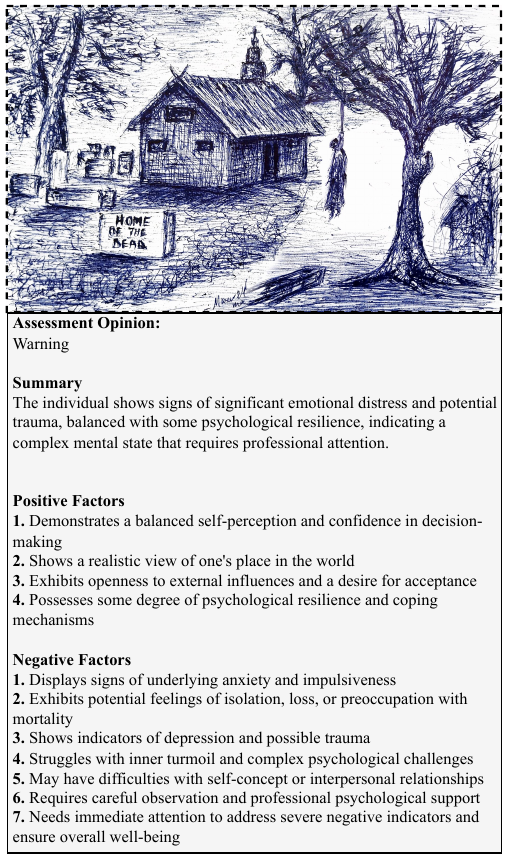}
    \caption{Case study of a Warning case.}
    \label{fig: case1}
\end{figure}

\subsection{Evaluation Results}
In this study, \method analyzes a total of 290 HTP drawings, classifying 90 cases (\textbf{31.03\%}) as \textbf{\textit{Warning}} and 200 cases (\textbf{68.97\%}) as \textbf{\textit{Observation}}.
The evaluation process involved psychological teachers reviewing their students' analysis reports generated by \method. Teachers assessed the consistency between these reports and their professional observations using a three-level scale: \textbf{High Consistency}, \textbf{Moderate Consistency}, and \textbf{Low Consistency}.

\begin{table}[!h]
\centering
\small
\resizebox{0.48\textwidth}{!}{%
\begin{tabular}{lccc}
\toprule
                & \textbf{Total (\%)} & \textbf{Warn. (\%)} & \textbf{Obs. (\%)} \\ \midrule
\textbf{\high}           & 71.03      & 58.89        & 76.50            \\
\textbf{\moderate} & 26.21      & 36.67        & 22.00            \\
\textbf{\low}       & 2.76       & 4.44         & 1.50             \\ \bottomrule
\end{tabular}%
}
\caption{Matching rates of results with teacher feedback.}
\label{tab: result}
\end{table}

\begin{figure}[t]
    \centering
    \includegraphics[width=\linewidth]{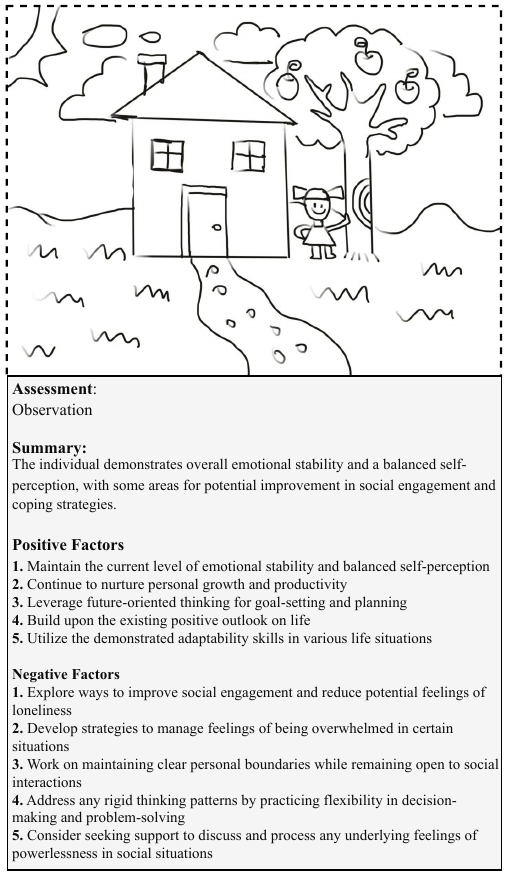}
    \caption{Case Study of an Observation case.}
    \label{fig: case2}
\end{figure}

Table \ref{tab: result} presents the evaluation results across different categories. Overall, the system achieved 71.03\% \high with teacher assessments. The Observation group demonstrated better performance with 76.50\% \high, while the Warning group showed 58.89\% \high. The higher consistency rate in the Observation group (76.50\%) compared to the Warning group (58.89\%) can be attributed to several factors. First, psychological issues often manifest in complex and diverse ways, requiring more nuanced professional judgment than typical psychological states. Second, \method is designed with a conservative strategy that tends to classify ambiguous cases as Warning, leading to more borderline cases in the Warning group. This conservative approach aligns with the system's primary goal of early screening, where false positives are preferred over missing potential psychological issues. It is important to note that \method is designed to assist, rather than replace, mental health professionals in early screening. The relatively lower consistency rate in the Warning group can be effectively addressed through subsequent professional evaluation, which is a necessary step in the psychological assessment process.

\subsection{Case Study of PsyDraw}
Figure \ref{fig: case1} presents a typical Warning case where dark themes and concerning elements (isolated house, hanging figure) are prominently displayed, reflecting potential psychological distress that requires professional attention. In contrast, Figure \ref{fig: case2} shows an Observation case featuring a balanced composition with positive elements (fruit-bearing tree, well-proportioned house, cheerful figure), indicating healthy psychological development.
All images used for case studies in this research were sourced from publicly accessible resources on the internet. While these brief summary reports provide quick insights, \method also enables mental health professionals to generate comprehensive reports with more detailed psychological analysis and interpretations\footnote{\url{\git}}.

\section{Conclusion}
This paper presents PsyDraw, a multi-agent system designed to assist mental health professionals in analyzing House-Tree-Person drawings for psychological screening of left-behind children. The system demonstrates promising results, achieving 71.03\% \high with professional assessments across 290 cases while identifying 31.03\% of cases requiring immediate attention (\textbf{\textit{Warning}}). This significant proportion of Warning cases highlights these communities' urgent need for psychological support. Through its two-stage analysis framework and specialized agents, PsyDraw effectively extracts psychological features and generates structured reports, particularly valuable in resource-limited rural areas. The system has been successfully deployed in several rural primary schools, assisting school counselors in early psychological screening. While not intended to replace professional judgment, PsyDraw offers a scalable solution to support early psychological screening where mental health resources are scarce. This research represents a step toward leveraging AI technologies to address critical mental health challenges in underserved communities while maintaining professional standards and ethical considerations.

\section{Ethics Statement}
This research prioritizes ethical considerations in both system development and deployment. The data collection process was conducted with explicit consent from participating schools, teachers, and students’ guardians. All HTP drawings were anonymized before analysis, with the development team having access only to the drawings themselves, without any personal identifiers or demographic information.

PsyDraw is designed specifically as an assistive tool for mental health professionals working in resource-limited areas, not as a replacement for professional psychiatric evaluation. The system includes built-in safeguards: when detecting severe psychological risk indicators (such as suicidal tendencies or violence), it immediately generates warning messages recommending professional intervention. This aligns with our principle of “safety first” in mental health screening.

To ensure responsible deployment, we emphasize the following guidelines:
\begin{itemize}
\item The system should only be used under the supervision of qualified mental health professionals or trained school counselors
\item Analysis results should be treated as preliminary screening indicators rather than definitive diagnoses
\item All data processing must comply with relevant privacy protection regulations
\item System access should be restricted to authorized personnel to protect sensitive information
\item Regular audits of the system’s performance and bias should be conducted to maintain reliability
\item Considering the potential risks of misuse, we have made the code open source but have not open-sourced the specific prompt templates. If users wish to utilize the full capabilities of \method, we encourage them to contact us via the GitHub repository.
\end{itemize}

We acknowledge that psychological assessment tools can significantly impact individuals’ lives. Therefore, we maintain a conservative approach in risk classification, preferring to flag potential concerns for professional review rather than potentially overlooking serious issues. This research aims to support, not substitute, professional mental health services, particularly in underserved rural areas with limited access to psychological resources.

This ethical framework will continue to guide future developments and deployments of the PsyDraw system, ensuring that technological advancement serves the best interests of vulnerable populations while maintaining professional standards and ethical principles in mental health care.

\section{Limitation}
The PsyDraw system is designed as an assistive tool for mental health professionals in resource-limited areas, primarily focusing on early screening of left-behind children's psychological well-being. While the system demonstrates promising results in supporting professional assessment, several limitations should be noted:

First, PsyDraw is specifically designed and validated for Chinese left-behind children in rural areas, and its effectiveness in other cultural contexts remains unverified. Second, as an MLLM-based system, it may exhibit inherent biases, and its performance stability needs further validation through longitudinal studies. Third, while the system can efficiently process drawings and generate structured reports, it should always work in conjunction with, rather than replace, professional judgment. The system serves as a preliminary screening tool to help mental health professionals identify potential cases requiring attention, optimizing the allocation of limited professional resources.

These limitations underscore PsyDraw's role as a supplementary tool in the psychological assessment workflow, particularly valuable in areas with scarce mental health resources. Future work should enhance the system's cultural adaptability and conduct long-term effectiveness studies.

\bibliography{custom}

\end{document}